\documentclass[conference]{IEEEtran}
\IEEEoverridecommandlockouts
\IEEEpubid{\makebox[\columnwidth]{979-8-3503-3607-8/23/\$31.00 ©2023 IEEE\hfill} \hspace{\columnsep}\makebox[\columnwidth]{ }}
\ifCLASSOPTIONcompsoc
  \usepackage[nocompress]{cite}
\else
  \usepackage{cite}

\fi
\ifCLASSINFOpdf
\else
\fi
\usepackage{amsmath,booktabs}
\usepackage{amssymb}
\usepackage{diagbox}
\usepackage{graphicx}
\usepackage[top=22.5mm, bottom=22.5mm, left=22.5mm, right=22.5mm]{geometry}
\usepackage{multirow}
\usepackage{color, colortbl}
\usepackage[breaklinks=true,letterpaper=true,colorlinks,bookmarks=false]{hyperref}
\usepackage[english]{babel}
\usepackage[dvipsnames]{xcolor}
\usepackage{makecell}

\begin{document}

%
\title{Face Morphing Attack Detection with Denoising Diffusion Probabilistic Models\thanks{Supported in parts by the ARRS research programme P2-0250 (B).}}


%

\author{\IEEEauthorblockA{Marija Ivanovska,
Vitomir \v{S}truc}
\IEEEauthorblockA{Faculty of Electrical Engineering, University of Ljubljana, Tr\v{z}a\v{s}ka cesta 25, Ljubljana, Slovenia\vspace{-3mm} }
}

\maketitle
\IEEEpubidadjcol

%

\begin{abstract}
Morphed face images have recently become a growing concern for existing face verification systems, as they are relatively easy to generate and can be used to impersonate someone's identity for various malicious purposes. 
Efficient Morphing Attack Detection (MAD) that generalizes well across different morphing techniques is, therefore, of paramount importance. Existing MAD techniques predominantly rely on discriminative models that learn from examples of bona fide and morphed images and, as a result, often 
exhibit sub-optimal generalization performance when confronted with unknown types of morphing attacks. 
To address this problem, we propose a novel, diffusion--based MAD method in this paper that learns only from the characteristics of bona fide images. Various forms of morphing attacks are then detected by our model as out-of-distribution samples. We perform rigorous experiments over four different datasets (CASIA-WebFace, FRLL-Morphs, FERET-Morphs and FRGC-Morphs) and compare the proposed solution to both discriminatively-trained and once-class MAD models. The experimental results show that our MAD model achieves highly competitive results on all considered datasets. 

\end{abstract}

\section{Introduction}
Automatic face recognition systems (FRSs) are widely used to verify a person's identity by matching the face image of an individual to the data enrolled in the system's database. 
While such systems are today widely deployed and highly accurate~\cite{grm2018strengths}, they are known to be prone to certain types of attacks with manipulated data, such as morphing attacks~\cite{Ferrara2016, Scherhag2017,SYN_MAD_2022}. Because face morphs are created by blending/morphing the facial appearances of at least two different people, a single morphed image can be utilized to falsely authenticate all individuals, whose face has been used during the morph-generation process.

With recent advancements in generative models and the availability of open-source morphing techniques, the generation of highly realistic, high--quality morphed face images has become an almost effortless process. The successful detection of \textit{face morphing attacks} is, hence, crucial for the prevention of illegal activities~\cite{Naser_PW_MAD_2021}. While significant progress has been achieved in morphing attack detection (MAD), the majority of existing solutions learn to detect morphed faces discriminatively, i.e., by analyzing and learning the differences between bona fide and morphed samples. 
Such techniques have been shown to be very accurate when evaluated on morphing techniques seen during training, but often fail to detect morphs created by unknown morphing attacks. Moreover, when evaluated on data from unknown sources, their accuracy is usually adversely affected by domain shifts. 

To address the generalization capabilities of MAD models, some researchers explored the use of the one-class models \cite{Damer2019_OC_MAD,Damer2022_OC_MAD_SPA,Ibsen2021_OC_MAD}, where only bona fide images are used in the training phase. Such models, are generally expected to generalize better to unseen morphs and are also at the heart of this work. Specifically, we propose in this paper a novel one-class MAD technique (MAD-DDPM) that exploits Denoising Diffusion Probabilistic Models (DDPMs) for the detection task. We evaluate the model in comprehensive experiments over multiple datasets and in comparison to both discriminatively-trained and one-class MAD competitors with promising results. 


\begin{figure}[t]
\begin{center}
\centering
  \includegraphics[width=0.9\linewidth]{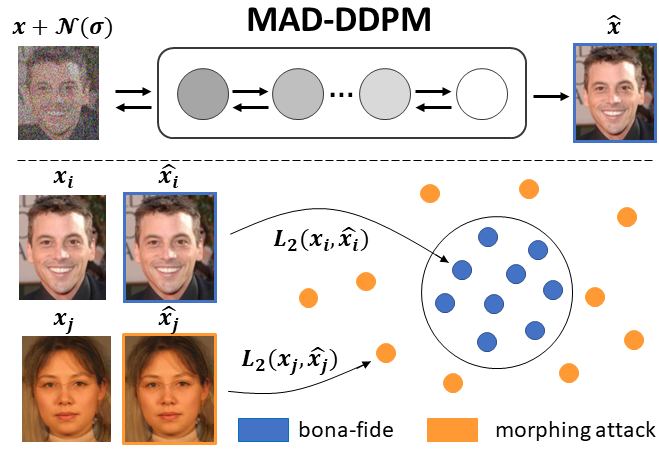}\vspace{-2mm}
\end{center}
\vspace{-10pt}
\caption{\textbf{Illustration of MAD-DDPM.} MAD-DDPM is a (reconstruct\-ion-based) one-class face morphing attack detection (MAD) model that uses a probabilistic (denoising) diffusion process to learn the distribution of bone-fide samples. At run-time, face morphs are detected based on the produced reconstruction error. Unlike the majority of competing MAD techniques, MAD-DDPM requires no attack examples during training.   \label{fig:MAD_DDPM_teaser}}
\end{figure}

\section{Related work}

In this section, we discuss background information and related work on morphing-attack-detection (MAD) and diffusion models to provide the necessary context for our research. For a more comprehensive coverage of these topics, the reader is referred to some of the excellent surveys available in  the literature, e.g., \cite{scherhag2019face,croitoru2022diffusion}.

\subsection{Morphing attack detection}
Existing morphing attack detection (MAD) models can in general be grouped into single-image (S-MAD) and differential (D-MAD) models. The first category of models examines facial morphs independently one from the other, while the latter are comparing manipulated samples to a reference. D-MADs are generally very accurate in closed-group problems, while S-MADs are predominantly used to detect attacks without prior knowledge of the subjects' identities. We limit the literature review in this section to S-MADs only, as they are most closely related to our work.

Regardless of the face morphing technique used, the generated morphs typically contain image irregularities, such noise, pixel discontinuities, distortions, spectrum discrepancies, and similar artifacts. With early MADs, such irregularities were often detected using hand-crafted techniques utilizing photo-response non-uniformity (PRNU) noise~\cite{Scherhag_PRNU_2019}, reflection analysis~\cite{Seibold_reflection_2018} or texture-based descriptors, such as LBP~\cite{Ojala_1996_LBP}, LPQ~\cite{Ojansivu_LPQ_2008} or SURF
~\cite{Makrushin_SURF_2019}. Although these methods yielded promising results, their generalization capabilities were shown to be limited~\cite{Damer2019_OC_MAD}.

More recent MADs take advantage of the capabilities of data--driven, deep-learning algorithms~\cite{SYN_MAD_2022}. Rag\-havendra \textit{et al.}~\cite{Raghavendra_transfer_2017} were amongst the first to propose transfer learning, with pretrained deep models for this task. In their work, attacks were detected with a simple, fully-connected binary classifier, fed with fused VGG19 and AlexNet features, pretrained on ImageNet. Wandzik \textit{et al.}~\cite{Wandzik_FRS_2018}, on the other hand, achieved high detection accuracy with features extracted with general-purpose face recognition systems (FRSs), fed to an SVM. Ramachandra \textit{et al.}~\cite{Ramachandra_Inception_2020} utilized Inception models in a similar manner, while Damer \textit{et al.}~\cite{Naser_PW_MAD_2021} argued that pixel-wise supervision, where each pixel is classified as a bona fide or a morphing attack, is superior, when used in addition to the binary, image-level objective. 
Recently, MixFaceNet~\cite{Boutros_MixFaceNet_2021} by Boutros \textit{et al.} achieved state-of-the-art results in different detection tasks, including face morphing detection~\cite{Damer_SMDD_2022}. This model represents a highly efficient architecture that captures different levels of attack cues through differently-sized convolutional kernels. 


Different from the supervised techniques discussed above, some authors have advocated the use of \textit{one-class learning} models trained on bone-fide samples only to improve the generalization capabilities of the MAD techniques. 
Damer \textit{et al.}~\cite{Damer2019_OC_MAD}, for example, were among the first to achieve significant performance generalization on unseen attacks with two different one--class methods, i.e. a one-class support vector machine (OCSVM) and an isolation forest (ISF). Similar generalization capabilities were later demonstrated in~\cite{Ibsen2021_OC_MAD}, where Ibsen \textit{et al.} explored the use of a Gaussian Mixture Model (GMM), Variational Autoencoder (VAE) and Single-Objective Generative
Adversarial Active Learning (SO-GAAL) in addition to an OCSVM. 
In a recent study, Fang \textit{et al.}~\cite{Damer2022_OC_MAD_SPA} proposed an unsupervised convolutional autoencoder, enhanced with a self-paced learning (SPL) algorithm. Here, the authors found 
that morphing attacks are easier to reconstruct in comparison to non-manipulated samples. 
The MAD-DDPM model, proposed in this paper, falls into the group of one-class learning models, but relies on a probabilistic diffusion process to learn the distribution of bona-fide face images.

\subsection{Diffusion models}
Denoising Diffusion Probabilistic Models (DDPMs) have recently been found to be exceptionally powerful models for various computer-vision tasks~\cite{croitoru2022diffusion,Saharia2022_palette, Vishal2023_T2V_DDPM}. DDPMs,  first introduced by Ho \textit{et al.}~\cite{Ho_DDPM_2020}, were shown to be able to generate high--quality images sampled from pure Gaussian noise. These methods learn to gradually add noise to training samples and to perform denoising, 
by iteratively maximizing the data likelihood. 
Although early models have shown impressive generative capabilities, their sampling techniques are time-consuming and often affect the image quality of the generated samples. 

Shortly after the initial release of DDPMs, Nichol \textit{et al.}~\cite{Nichol2021_improved_ddpm} proposed an improved optimization criterion that significantly sped up the noise removal, while maintaining the quality of the generated data. Song \textit{et al.}~\cite{song2020_DDIM} proposed their own solution for faster sampling and easier deployment of the diffusion process. Dhariwal \textit{et al.}~\cite{Dhariwal2021_DDPMbeatGANs} built on these findings and  showed that DDPMs can outperform GANs on image synthesis. 
In a recent study, Karras \textit{et al.}~\cite{Karras2022edm} explored different approaches for image generation with diffusion and provided guidelines related to the architectural design and the optimization strategy of DDPMs. Rombach \textit{et al.}~\cite{Rombach_2022_CVPR_latent_dif} successfully reduced the complexity of the diffusion models, by implementing the diffusion process in the latent space of a pretrained autoencoder with minimal degradation in image quality.

\begin{figure*}[t]
\begin{center}
\centering
  \includegraphics[width=0.9\textwidth]{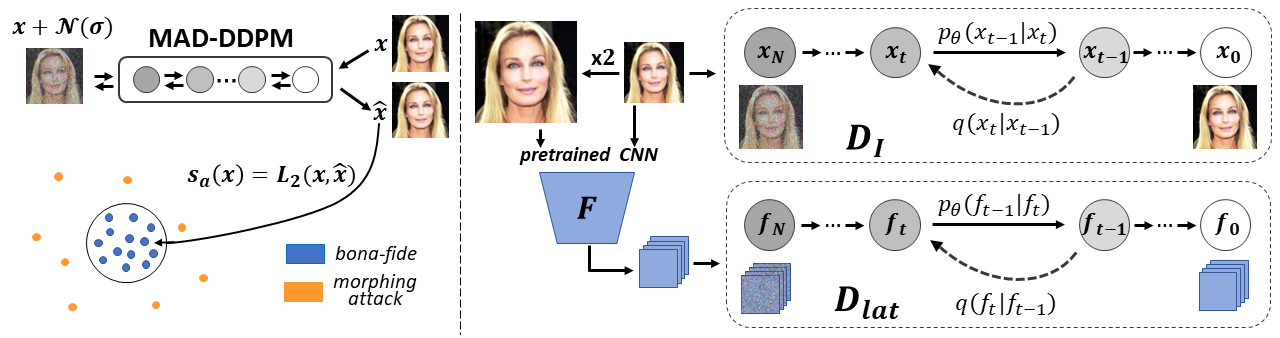}
\end{center}
\vspace{-12pt}
\caption{\textbf{High-level overview of the proposed MAD-DDPM model.} MAD-DDPM is a one-class learning model that uses a reconstruction-based measure to determine whether the input images are bona fide or face morphs (shown on the left). At the core of the technique is a two-branch reconstruction procedure that uses denoising diffusion probabilistic models (DDPMs) learned over only bona-fide samples as the basis for the detection tasks (shown on the right). Here, the first branch models the distribution on bona-fide samples directly in the pixel-space (for low-level artifact detection), while the second captures the distribution of higher-level features extracted with a pretreind CNN $F$.  \label{fig:MAD_DDPM}\vspace{-1mm}}
\end{figure*}

Although DDPMs were primarily developed for the generation of new data, they have also been adapted to one-class-learning algorithms. Wolleb \textit{et al.}~\cite{wolleb2022_DDPM_medical}, for example, trained an image-to-image diffusion model, that learned to reconstruct medical images of healthy subjects through the iterative denoising process. 
A similar technique, proposed by Wyatt \textit{et al.}~\cite{Wyatt2022_AnoDDPM}, used Simplex noise, instead of the common Gaussian noise. In contract, Pinaya \textit{et al.}~\cite{Pinaya2022_AD_latent_MICCAI} detected anomalies by utilizing diffusion models in the latent space, where the noise reversal is much more efficient.

\section{Methodology}

A considerable cross-section of existing MAD techniques uses discriminatively trained models for the detection of facial morphs and, as a result, often struggles with the generalization to unseen morphing attacks. In this section, we propose a novel MAD model, MAD-DDPM, that is trained  with bona-fide samples only and is, therefore, expected to generalize better to various (unknown, unseen) types of morphing attacks.

\subsection{Overview of MAD-DDPM}

A high-level overview of the proposed MAD-DDPM model is presented in Figure~\ref{fig:MAD_DDPM}. The model follows the (self-supervised, one-class) reconstruction-based framework to anomaly detection~\cite{Ganomaly, ZAVRTANIK2021107706}, where a generative model is learned to reconstruct so-called \textit{normal} training data, i.e., bona-fide face images, frome noisy inputs. Because \textit{anomalies} (face morphs in our case) deviate from the distribution of the normal samples, they are expected to generate (comparably) larger reconstruction errors. Consequently, these errors can be used to determine whether the given data sample is normal (bona-fide) or anomalous (morph), as illustrated on the left of Figure~\ref{fig:MAD_DDPM}. 

While different generative models have been used in the literature for reconstruction-based anomaly detection (e.g., autoencoders, GANs, etc.), they were often observed to generalize too well beyond the training data, leading to comparable reconstructions for both normal and anomalous data. For the MAD-DDPM we, therefore, design a powerful reconstruction procedure that: $(i)$ results in larger differences in the reconstructions of bona-fide and morphed images than competing (one-class) MAD solutions, and $(ii)$ consequently results in better performance. The procedure is based on Denoising Diffusion Probabilistic Models (DDPMs) and the following considerations:
\begin{itemize}
\item \textbf{Complementary data representation:} The reconstruction task is learned over two data representation, i.e., $(i)$ the pixel space, where the goal is to model image-level (bona-fide) facial characteristics and to facilitate the detection of low-level image artifacts, and $(ii)$ a feature space that captures higher-level semantic cues of the training data, enabling the detection of potentially more abstract data irregularities.
\item \textbf{Compact distribution modelling:} To ensure the generative models do not generalize too well beyond the data used for learning, efficient modeling techniques are needed that result in compact distributions of the training data. In MAD-DDPM, we model the data distribution of the bona-fide samples using a probabilistic denoising diffusion process across two data representations, which allows us to efficiently capture the characteristics of the bona-fide samples in a compact manner. This leads to highly competitive MAD performance, as demonstrated in Section~\ref{Sec: Results}.  
\end{itemize}

In the following sections, we present the theoretical background behind DDPMs, discuss the design of the MAD-DDPM reconstruction procedure, and elaborate on the detection-score computation step.

\subsection{Denoising Diffusion Probabilistic Models (DDPMs)}

DDPMs are likelihood--based generative methods, that learn to model a given data distribution $p_{data}(\mathbf{x})$ with standard deviation $\sigma_{data}$ by employing a two--stage approach~\cite{Ho_DDPM_2020}. In the first stage, a forward diffusion process is applied to the data $\mathbf{x}_0\sim p_{data}(\mathbf{x})$, by gradually corrupting the sample $x_0$ with Gaussian noise $\mathcal{N}(0, \sigma^2\mathbf{I})$. The noising technique results in a noisy sample $x_N$ and represents a non--homogeneous Markov chain:
\begin{equation}\label{eq:forward}
    q(\mathbf{x}_t|\mathbf{x}_{t-1}) = \mathcal{N}(\mathbf{x}_t|\mathbf{x}_{t-1}\sqrt{1-\beta_t}, \beta_t\mathbf{I})
\end{equation}
where $t$ is the time step from a predefined time sequence $\{t_0, t_1,... t_N\}$, while $\beta_t=\sigma_t^2$ defines the amount of noise added at each step and its value is determined by a variance schedule. Following the recommendations from~\cite{Karras2022edm}, we implement a linear variance schedule, found to work best in terms of sampling speed and generated data quality. The forward process defined with Eq.~(\ref{eq:forward}) enables fast sampling of $\mathbf{x}_t$ at any time step $t$:
$$
q(\mathbf{x}_t|\mathbf{x}_0) = \mathcal{N}(\mathbf{x}_t; \sqrt{\bar{\alpha_t}}\mathbf{x}_0, (1-\bar{\alpha_t})\mathbf{I})
$$
where $\bar{\alpha_t}=1-\beta_t$ and $\bar{\alpha_t}=\Pi_{i=1}^t \alpha_i$. In the second stage, a generative model parametrized by $\theta$ performs sequential denoising of $\mathbf{x}_N$ according to:
\begin{equation}\label{eq:reverse}
    p_{\theta}(\mathbf{x}_{t-1}|\mathbf{x}_t, \mathbf{x}_0)=\mathcal{N}(\mathbf{x}_{t-1}|\tilde{\mu}_{t}(\mathbf{x}_t, \mathbf{x}_0), \tilde{\beta_t}\mathbf{I})
\end{equation}
where $t=t_N, t_{N-1},... t_0$, $\tilde{\mu}_{t}(\mathbf{x}_t, \mathbf{x}_0)=\frac{\sqrt{\bar{\alpha}_{t-1}}\beta_t}{1-\bar{\alpha}_t}\mathbf{x}_0+\frac{\sqrt{\bar{\alpha}_t}(1-\bar{\alpha}_{t-1})}{1-\bar{\alpha}_t}x_t$ and $\tilde{\beta_t}=\frac{1-\bar{\alpha}_{t-1}}{1-\bar{\alpha}_{t}}\beta_t$. The mean function $\tilde{\mu}_{t}$ is optimized by an aproximator $\mathcal{D}_\theta(\mathbf{x}, \sigma)$, trained to minimize the expected $L_2$ denoising error:
\begin{equation}
    \mathcal{L} = \mathbb{E}_{\mathbf{x}\sim p_{data}}\mathbb{E}_{\mathbf{n}\sim \mathcal{N}(0, \sigma^2\mathbf{I})} ||\mathcal{D}(\mathbf{x}+\mathbf{n}; \sigma)-\mathbf{x}||_2^2
\end{equation}
where $\mathcal{D}_\theta(\mathbf{x}, \sigma)$ is a neural network. For MAD-DDPM, an unconditional U-Net~\cite{PixelCNN_ICLR2017_Kingma} architecture, originally proposed in~\cite{Ho_DDPM_2020}, is selected for the implementation of this network. For efficiency reasons, we leverage the recently published DPM-Solver~\cite{Lu_NIPS2022_DPM_solver}, a dedicated high-order
solver for diffusion ordinary differential equations (ODEs).

\subsection{Reconstruction and Detection-Score Computation}

MAD-DDPM uses a two-branch reconstruction procedure to model the distribution of the bona fide samples, as shown in Figure~\ref{fig:MAD_DDPM}. The first DDPM branch, $\mathcal{D}_I$, is modeling the distribution of bona fide face images in the pixel space. The second DDPM branch, $\mathcal{D}_{lat}$, operates in the feature space of a pretrained convolutional network $F$, that extracts high-level image representations over two different scales. Here, the calculated feature maps are concatenated before feeding them to the dedicated diffusion model. Each DDPM branch of the model is learned independently of the other to reduce the computational effort and reduce cross-talk and interactions between low-level image characteristics and higher-level semantic cues.   

During run-time, the probability of an image $\mathbf{x}_n$ to be a morphing attack is quantified using the score $s_a$, calculated by summing up the reconstruction errors of the two diffusion branches, i.e.:
\begin{equation}\label{eq:attack_score}
    s_a(\mathbf{x}_n) = \mathcal{D}_I(\mathbf{x}_n+\mathbf{n}_I; \sigma_I)+\mathcal{D}_{lat}(F(\mathbf{x}_n)+\mathbf{n}_F; \sigma_{F})
\end{equation}

Because our main goal is to detect face morphing artifacts, MAD-DDPM performs the iterative noising with a relatively low $\sigma_{max}$, which leads to moderately noised samples. In contrast to existing generative DDPMs, our model is, therefore, unable to generate new samples directly from noise. Instead, it is conditioned on the noisy input $\mathbf{x}_n+\mathbf{n}_I$ and aims to recover information that has been obscured during the forward noising process. 

\section{Experimental Setup}

\subsection{Datasets}
We primarily use four publicly available datasets for the experiments: CASIA-WebFace~\cite{CASIA-WebFace}, FERET-Morphs, FRLL-Morphs and FRGC-Morphs~\cite{Sarkar2020_morphed_data}. Images from all datasets are first preprocessed by RetinaFace~\cite{RetinaFace} to localize the facial areas. Next, these areas are cropped with a margin equal to $5\%$ of the bounding box height. With this strategy, we ensure, that the cropped images include pixels surrounding the face area, as this is where a considerable amount of morphing artifacts is typically located. Finally, the cropped images are resized to $224\times224$ pixels and fed to the MAD model. The training of the model is performed in a one--class learning manner, with bona fide images only. In the testing phase, we use three different datasets consisting of both, bona fide and morphed images.

\textbf{Training data.} The MAD models are trained on CASIA-WebFace \cite{CASIA-WebFace}, a large-scale dataset  used commonly for  face verification and identification tasks. The dataset consists of $494.414$ face images of $10.575$ unique subjects, collected from the internet. The dataset was designed to include a wide variety of face poses and expressions, captured under different illumination settings and with different image resolutions.
\begin{figure}[t]
\begin{center}
\centering
  \includegraphics[width=0.9\linewidth]{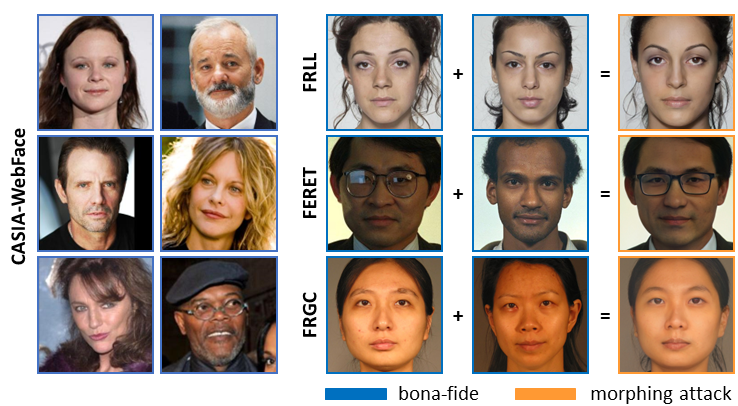}
\end{center}
\vspace{-15pt}
\caption{\textbf{Sample images from the datasets used for the evaluation.} The figure shows bona fide training images from CASIA-WebFace~\cite{CASIA-WebFace} (left) and morphed images and their respective bona fide source faces from FERET-Morphs, FRLL-Morphs and FRGC-Morphs~\cite{Sarkar2020_morphed_data} (right). \label{fig:datasets}\vspace{-2mm}}
\end{figure}

\textbf{Testing data.}
Testing is done on three common morphing datasets proposed by Sarkar \textit{et al.} in~\cite{Sarkar2020_morphed_data}, i.e. FRLL-
Morphs, FERET-Morphs and FRGC-Morphs. All morphed face images were created by merging bona fide samples from their respective source datasets, i.e. FRLL~\cite{debruine_jones_2017, Neubert_FRLL}, FERET~\cite{PHILLIPS_FERET} and FRGC~\cite{Phillips_FRGC}. For the generation of the landmark--based morphs, the authors used OpenCV and FaceMorpher, 
while deep--learning--based morphs were generated with StyleGAN2. Additionally, image samples from FRLL, have been used as a source for morph generation with two more morphing methods, AMSL\cite{Neubert_FRLL} and Webmorph.
The  characteristics of the datasets 
are given in Table~\ref{tab1} and a few examples are presented in Figure~\ref{fig:datasets}. 

\textbf{Training data for supervised MADs.} MAD-DDPM is also evaluated against selected discriminatively trained MAD methods, learned on morphs from $3$ datasets not used for our evaluations, i.e. LMA-DRD~\cite{Naser_PW_MAD_2021}, MorGAN~\cite{MorGAN} and SMDD~\cite{Damer_SMDD_2022}. LMA-DRD morphs are generated with OpenCV. Digital morphs are labeled with D, while re--digitalized (printed then scanned) morphs are labeled with PS. MorGAN also consists of two types of moprhs: LMA, generated with OpenCV and deep learning-based morphs, generated with a GAN model. SMDD, on the other hand, contains synthetically generated data, where both, bona fide and attack samples are created with StyleGAN2.

\begin{table}[t]
\caption{Number of bona fide images (BF) and morphing attacks (MA) in each test dataset. The morphs are generated by $5$ morphing methods, i.e. OpenCV (OCV), FaceMorpher (FM), StyleGAN (SG), AMSL, Webmorph (WM).\vspace{-4mm}}
\label{tab1}
\begin{center}
\resizebox{\columnwidth}{!}{%
\begin{tabular}{  l | c | c | c | c | c | c | c }
\toprule 
  \textbf{Dataset} & \textbf{Image size} & \textbf{BF} & \textbf{OCV} & \textbf{FM} & \textbf{SG} & \textbf{AMSL} & \textbf{WM} \\ 
\midrule  
  FRLL-M & $1350\times1350$ & $204$ & $1221$ & 1222 & 1222 & 2175 & 1221 \\ 
  FERET-M & $512\times768$ & $1.413$ & 529 & 529 & 529 & / & / \\
  FRGC-M & $227\times277$ & $3.167$ & 964 & 964 & 964 & / & / \\
\bottomrule
\end{tabular}}\vspace{-4mm} 
\end{center}
\end{table}


\subsection{Evaluation metrics}
The model evaluation follows the testing protocol proposed in~\cite{Damer2022_OC_MAD_SPA}. Based on morphing attack scores, we first calculate the proportion of attack samples misclassified as bona fide, i.e. the Attack Presentation Classification Error Rate (APCER). We also calculate the proportion of bona fide samples misclassified as attacks, i.e. the Bona fide Presentation Classification Error Rate (BPCER). The overall detection accuracy is then reported in terms of the Equal Error Rate (EER), where APCER equals BPCER.

\subsection{Implementation details}

The input to MAD-DDPM consists of RGB images and the corresponding feature maps extracted with a WideResNet50 \cite{WideResNet} (model $F$), pretrained on ImageNet. The feature extraction is performed on two different scales, to better capture differently sized patterns. First, images of size $224\times224$ are fed to the WideResnet to calculate feature maps of size $1024\times14\times14$. Next, each RGB image is resized and split into $4$ non--overlapping patches, that are passed through WideResNet, to get $4$ additional feature maps. The  DDPM branch, labeled as $\mathcal{D_I}$ (Figure~\ref{fig:MAD_DDPM}), is then optimized on raw RGB images with $\sigma_{max}=8$, while $\mathcal{D}_{lat}$ is trained on the concatenated feature maps, with $\sigma_{max}=2$. The $\sigma_{max}$ values were determined based on preliminary experiments. 
The DDPMs in the proposed model are optimized with AdamW~\cite{AdamW_ICLR2019}. The learning rate is set to $0.0001$, $\beta_1$ and $\beta_2$ to $0.95$ and $0.999$, respectively, while the weight decay is set to $0.001$. 

MAD-DDPM is implemented in Python 3.8 with PyTorch 1.9 and CUDA 11.7. Experiments were run on a single NVIDIA GeForce RTX 3090, where MAD-DDPM required around $0.6s$ to perform the MAD procedure for a single image on average. The source code od MAD-DDPM is available at https://github.com/MIvanovska/MAD-DDPM.

\section{Results}\label{Sec: Results}

\textbf{Comparison to One-Class Competitors.} We first compare MAD-DDPM to the current state-of-the-art (SOTA) one-class  SPL-MAD approach~\cite{Damer2022_OC_MAD_SPA}. The results in Table~\ref{tab:quantitative_results} show that MAD-DDPM achieves very competitive results on FRLL-Morphs. In the detection of StyleGAN morphing attacks, it outperforms SPL-MAD by over $5\%$ in terms of EER, while producing comparable results on the remaining morphs. On the other two datasets, 
MAD-DDPM consistently outperforms SPL-MAD across all types of morphing attacks. Overall, MAD-DDPM achieves an average EER of $16.88\%$, outperforming the current one-class SOTA method by a margin of over $4\%$. 

\begin{table}[t]
\caption{Comparison of MAD-DDPM and the current SOTA one-class SPL-MAD approach in terms of EER (\%).} 
\vspace{-10pt}
\label{tab:quantitative_results}
\smallskip
\begin{center}
\resizebox{0.99\columnwidth}{!}{%
\begin{tabular}{  c | l | c | c  }
\toprule
  \multirow{2}{*}{\textbf{Dataset}} & \multirow{2}{*}{\textbf{Morphing methods}} & \multirow{2}{*}{\textbf{SPL-MAD~\cite{Damer2022_OC_MAD_SPA}}} & \multirow{2}{*}{\textbf{MAD-DDPM (Ours)}} \\
   &  &  & \\ \midrule
  \multirow{5}{*}{FRLL-M} & OpenCV & $3.63$ & $\mathbf{3.55}$\\
& FaceMorpher & $\mathbf{2.98}$ & $4.04$\\
  & StyleGAN2 & $15.14$ & $\mathbf{10.96}$\\
  & WebMorph & $\mathbf{12.29}$ & $14.49$\\
  & AMSL & $\mathbf{11.22}$ & $11.67$ \\\midrule 
  \multirow{3}{*}{FERET-M} & OpenCV & $32.13$ & $\mathbf{30.81}$\\
& FaceMorpher & $27.69$ & $\mathbf{25.14}$\\
  & StyleGAN2 & $32.57$ & $\mathbf{23.25}$\\\midrule
  \multirow{3}{*}{FRGC-M} & OpenCV & $36.11$ & $\mathbf{27.17}$\\
& FaceMorpher & $23.99$ & $\mathbf{23.23}$\\
  & StyleGAN2 & $36.79$ & $\mathbf{11.41}$\\\midrule
  \multicolumn{2}{c|}{Average performance} & $21.32$ & $\mathbf{16.88}$ \\
\bottomrule
\end{tabular}}\vspace{-4mm}
\end{center}
\end{table}

\textbf{Comparison to Discriminative MAD models.} Similarly to~\cite{Damer2022_OC_MAD_SPA}, we also compare MAD-DDPM to SOTA discriminative MAD techniques in Table~\ref{tab:comparison_results}, i.e., MixFaceNet~\cite{Boutros_MixFaceNet_2021}, PW-MAD~\cite{Naser_PW_MAD_2021} and Inception~\cite{Ramachandra_Inception_2020}. The discriminative models are learned in a two-class setting, where a different set of morphing attacks is chosen in each training session. Although the best EER in individual categories of morphing attacks is achieved by the discriminative MADs, none of the trained discriminative models shows consistently superior results across different datasets and morphing attack types. Moreover, the average morphing attack detection performance is by far the highest for MAD-DDPM with an average EER of $16.88\%$. Based on these results, we conclude that our one-class MAD-DDPM approach demonstrates strong generalization capabilities. 

\begin{table*}[t]
\caption{Comparison of MAD-DDPM and discriminative SOTA MADs in terms of EER (\%). 
\vspace{-6mm}}
\label{tab:comparison_results}
\begin{center}
\resizebox{\textwidth}{!}{%
\begin{tabular}{  c | l | c | c | c | c | c | c | c | c | c | c | c | c | c | c | c | c  }
\toprule  
\multicolumn{2}{l}{\textbf{MAD type}} & \multicolumn{15}{|c|}{\textbf{Discriminativelly trained}} & \textbf{One-class}\\ \cmidrule{1-18}
\multicolumn{2}{l|}{\textbf{MAD model}} & \multicolumn{5}{c|}{\textbf{MixFaceNet}~\cite{Boutros_MixFaceNet_2021}} & \multicolumn{5}{c|}{\textbf{PW-MAD}~\cite{Naser_PW_MAD_2021}} & \multicolumn{5}{c|}{\textbf{Inception}~\cite{Ramachandra_Inception_2020}} & \multirow{2}{*}{\textbf{\makecell{MAD-DDPM \\ (Ours)}}}\\ \cmidrule{1-17}
\multicolumn{2}{l|}{\backslashbox{Test data}{Train data}} & D & PS & LMA & GAN & SMDD & D & PS & LMA & GAN & SMDD & D & PS & LMA & GAN & SMDD & \\\midrule
\multirow{5}{*}{FRLL-M} & OpenCV & $8.82$ & $13.22$ & $8.91$ & $17.66$ & $4.39$ & $17.33$ & $15.69$ & $13.96$ & $45.59$ & $\mathbf{2.42}$ & $13.72$ & $10.76$ & $6.86$ & $55.89$ & $5.38$ & $3.55$ \\
& FaceMorpher & $7.80$ & $10.97$ & $7.34$ & $15.65$ & $3.87$ & $13.88$ & $15.14$ & $10.92$ & $44.57$ & $\mathbf{2.20}$ & $16.62$ & $15.81$ & $6.32$ & $66.14$ & $3.17$ & $4.04$ \\
  & StyleGAN2 & $20.07$ & $15.29$ & $13.41$ & $23.51$ & $\mathbf{8.89}$ & $29.97$ & $27.64$ & $18.11$ & $48.53$ & $16.64$ & $37.24$ & $19.58$ & $20.56$ & $55.03$ & $11.37$ & $10.96$ \\
  & WebMorph & $25.97$ & $29.04$ & $20.61$ & $30.39$ & $12.35$ & $33.78$ & $28.51$ & $35.75$ & $52.43$ & $16.65$ & $57.38$ & $58.32$ & $30.88$ & $77.42$ & $\mathbf{9.86}$ & $14.49$ \\
  & AMSL & $24.53$ & $27.59$ & $19.24$ & $30.03$ & $15.18$ & $36.25$ & $32.95$ & $34.38$ & $48.52$ & $15.18$ & $49.02$ & $61.44$ & $\mathbf{9.80}$ & $86.49$ & $10.79$ & $11.67$ \\\midrule

\multirow{3}{*}{FERET-M} & OpenCV & $28.12$ & $32.19$ & $31.57$ & $33.86$ & $31.74$ & $37.27$ & $45.29$ & $34.27$ & $43.11$ & $39.93$ & $\mathbf{6.39}$ & $7.23$ & $42.12$ & $13.62$ & $59.32$ & $30.81$ \\
& FaceMorpher & $22.57$ & $29.48$ & $27.90$ & $31.81$ & $23.69$ & $35.16$ & $44.30$ & $28.24$ & $40.40$ & $29.41$ & $\mathbf{5.17}$ & $6.91$ & $36.53$ & $18.36$ & $46.94$ & $25.14$ \\
& StyleGAN2 & $29.57$ & $29.02$ & $35.46$ & $39.41$ & $39.85$ & $44.25$ & $45.30$ & $29.70$ & $42.47$ & $47.20$ & $9.03$ & $\mathbf{7.12}$ & $35.29$ & $15.09$ & $60.05$ & $23.25$ \\\midrule

\multirow{3}{*}{FRGC-M} & OpenCV & $23.81$ & $25.04$ & $31.62$ & $21.11$ & $20.67$ & $57.06$ & $48.60$ & $29.74$ & $53.55$ & $26.45$ & $34.32$ & $\mathbf{13.65}$ & $36.17$ & $59.66$ & $19.63$ & $27.17$ \\
& FaceMorpher & $22.83$ & $23.54$ & $29.38$ & $19.98$ & $18.10$ & $56.00$ & $50.70$ & $30.49$ & $51.61$ & $23.40$ & $34.96$ & $19.71$ & $35.10$ & $56.91$ & $\mathbf{16.06}$ & $23.23$ \\
& StyleGAN2 & $32.71$ & $28.68$ & $21.70$ & $21.95$ & $11.62$ & $37.38$ & $38.42$ & $16.43$ & $26.62$ & $14.32$ & $41.14$ & $25.85$ & $36.19$ & $47.03$ & $15.26$ & $\mathbf{11.41}$ \\\midrule
\multicolumn{2}{c|}{\textbf{Average performance}} & $22.43$ & $24.01$ & $22.47$ & $26.03$ & $17.30$ & $36.21$ & $35.69$ & $25.64$ & $45.22$ & $21.25$ & $27.73$ & $22.40$ & $26.89$ & $50.15$ & $23.44$ & $\mathbf{16.88}$ \\ \bottomrule
\multicolumn{18}{l}{$^*$D: LMA-DRD (D), PS: LMA-DRD (PS), LMA: MorGAN (LMA), GAN: MorGAN (GAN)} \vspace{-5mm}

\end{tabular}}
\end{center}
\end{table*}

\textbf{Ablation study.}
MAD-DDPM is trained on three different data sources, i.e. RGB images (I) and feature maps from two different image scales (S1 and S2). The contribution of each data source is investigated in an ablation study, where we train three independent DDPMs, one for each data source. A separate DDPM, is trained with concatenated CNN features of both scales. As can be seen from Table~\ref{tab:ablation_study}, among all ablated models, the highest detection accuracy is achieved by the DDPM trained on RGB images. We hypothesize, that due to the nature of DDPMs, such approach efficiently detects high-frequency components representing image artifacts induced by the morphing techniques. Conversely, the extracted features encode high-level semantics that are comparably less informative (yet still important) for the morphing detection task. They do however consistently boost the detection performance in all test datasets. The complete MAD-DDPM model outperforms all ablated models with an average EER of $16.88\%$. 

\begin{table}[t]
\caption{Results of the ablation study.\vspace{-3mm}} 
\label{tab:ablation_study}
\begin{center}
\resizebox{0.95\columnwidth}{!}{%
\begin{tabular}{  c | l | c | c | c | c | c }
\toprule
  \multirow{2}{*}{\textbf{Dataset}} & \multirow{2}{*}{\textbf{Morphs}} & \multicolumn{5}{c}{\textbf{Data source}} \\ \cmidrule{3-7}
  & & \textbf{I} & \textbf{S1} & \textbf{S2} & \textbf{S1 + S2} & \textbf{I + S1 + S2} \\\midrule
  \multirow{5}{*}{FRLL-M} & OpenCV & $6.55$ & $32.02$ & $30.88$ & $24.65$ & $\mathbf{3.55}$\\\cmidrule{2-7}
& FaceMorpher & $4.21$ & $26.63$ & $21.91$ & $20.18$ & $\mathbf{4.04}$\\
  & StyleGAN2 & $12.11$ & $26.60$ & $20.70$ & $17.51$ & $\mathbf{10.96}$\\
  & WebMorph & $14.82$ & $32.92$ & $41.85$ & $38.74$ &$\mathbf{14.49}$\\
  & AMSL & $12.02$ & $34.53$ & $40.97$ & $34.89$ & $\mathbf{11.67}$ \\\midrule
  \multirow{3}{*}{FERET-M} & OpenCV & $31.38$ & $41.78$ & $40.07$ & $37.61$ & $\mathbf{30.81}$\\
& FaceMorpher & $25.52$ & $35.73$ & $32.70$ & $31.56$ & $\mathbf{25.14}$\\
  & StyleGAN2 & $34.59$ & $32.70$ & $23.44$ & $\mathbf{23.15}$ &$23.25$\\\midrule
  \multirow{3}{*}{FRGC-M} & OpenCV & $28.42$ & $28.53$ & $28.01$ & $\mathbf{25.00}$ & $27.17$\\
& FaceMorpher & $24.69$ & $24.38$ & $25.00$ & $22.71$ & $\mathbf{23.23}$\\
  & StyleGAN2 & $12.34$ & $24.69$ & $15.35$ & $13.09$ & $\mathbf{11.41}$\\\midrule
  \multicolumn{2}{c|}{\textbf{Average performance}} & $18.79$ & $30.96$ & $29.17$ & $26.28$ & $\mathbf{16.88}$ \\
\bottomrule
\end{tabular}}
\end{center}
\vspace{-5mm}
\end{table}

\section{Conclusion}
We presented a one-class model for morphing attack detection (MAD) that relies on denoising diffusion probabilistic models (DDPM). In comprehensive experiments, the model was shown to result in highly competitive performance on multiple datasets. As part of our future work, we plan to incorporate additional proxy task into the proposed model to further improve results. 





%

\bibliographystyle{ieee}
\bibliography{bibliography}




\end{document}